\documentclass[conference]{IEEEtran}

\makeatletter

\newcommand{\linebreakand}{%
  \end{@IEEEauthorhalign}
  \hfill\mbox{}\par
  \mbox{}\hfill\begin{@IEEEauthorhalign}
}

\def\ps@IEEEtitlepagestyle{%
  \def\@oddfoot{\mycopyrightnotice}%
  \def\@evenfoot{}%
}
\def\mycopyrightnotice{%
  {\footnotesize 979-8-3503-2781-6 /23/\$31.00~\copyright~2023 IEEE\hfill}
  \gdef\mycopyrightnotice{}
}

\usepackage{blindtext}
\usepackage{eso-pic}
\IEEEoverridecommandlockouts
\usepackage{cite}
\usepackage{amsmath,amssymb,amsfonts}
\usepackage{algorithmic}
\usepackage{graphicx}
\usepackage{textcomp}
\usepackage{xcolor}
\def\BibTeX{{\rm B\kern-.05em{\sc i\kern-.025em b}\kern-.08em
    T\kern-.1667em\lower.7ex\hbox{E}\kern-.125emX}}
    
\usepackage{eso-pic}
\newcommand\AtPageUpperMyright[1]{\AtPageUpperLeft{%
 \put(\LenToUnit{0.17\paperwidth},\LenToUnit{-2cm}){%
     \parbox{0.9\textwidth}{\raggedleft\fontsize{8}{11}\selectfont #1}}%
 }}%
\newcommand{\conf}[1]{%
\AddToShipoutPictureBG*{%
\AtPageUpperMyright{#1}
}
}    
\usepackage{booktabs}
\usepackage{listings}
\usepackage{scalerel}
\usepackage[binary-units]{siunitx}
\usepackage{tikz, pgfplots}
  \usetikzlibrary{svg.path}

\makeatletter
\let\NAT@parse\undefined
\makeatother
\usepackage{hyperref}

\newcommand\httpsurl[1]{%
  \href{https://#1}{\nolinkurl{#1}}%
}

\definecolor{orcidlogocol}{HTML}{A6CE39}
\tikzset{
  orcidlogo/.pic={
    \fill[orcidlogocol] svg{M256,128c0,70.7-57.3,128-128,128C57.3,256,0,198.7,0,128C0,57.3,57.3,0,128,0C198.7,0,256,57.3,256,128z};
    \fill[white] svg{M86.3,186.2H70.9V79.1h15.4v48.4V186.2z}
                 svg{M108.9,79.1h41.6c39.6,0,57,28.3,57,53.6c0,27.5-21.5,53.6-56.8,53.6h-41.8V79.1z M124.3,172.4h24.5c34.9,0,42.9-26.5,42.9-39.7c0-21.5-13.7-39.7-43.7-39.7h-23.7V172.4z}
                 svg{M88.7,56.8c0,5.5-4.5,10.1-10.1,10.1c-5.6,0-10.1-4.6-10.1-10.1c0-5.6,4.5-10.1,10.1-10.1C84.2,46.7,88.7,51.3,88.7,56.8z};
  }
}
\newcommand\orcidicon[1]{\href{https://orcid.org/#1}{\mbox{\scalerel*{
\begin{tikzpicture}[yscale=-1,transform shape]
\pic{orcidlogo};
\end{tikzpicture}
}{|}}}}

\newcommand\copyrighttext{%
    \scriptsize \copyright{ }2023 IEEE. Personal use of this material is permitted. Permission from IEEE must be obtained for all other uses, in any current or future media, including reprinting/republishing this material for advertising or promotional purposes, creating new collective works, for resale or redistribution to servers or lists, or reuse of any copyrighted component of this work in other works.}
\newcommand\copyrightnotice{%
    \begin{tikzpicture}[remember picture,overlay]
    \node[anchor=south,yshift=10pt,xshift=7pt] at (current page.south) {\parbox{\dimexpr\textwidth-\fboxsep-\fboxrule\relax}{\copyrighttext}};
    \end{tikzpicture}%
}

\begin{document}
\title{\vspace*{1cm} Enhancing Lidar-based Object Detection in Adverse Weather using Offset Sequences in Time\\
\thanks{\textsuperscript{\textdagger}These authors contributed equally.}
}

\author{
\IEEEauthorblockN{Raphael van Kempen\textsuperscript{\textdagger\orcidicon{0000-0001-5017-7494}}}
\IEEEauthorblockA{\textit{Institute for Automotive Engineering} \\
\textit{RWTH Aachen University}\\
Aachen, Germany \\
raphael.vankempen@ika.rwth-aachen.de}
\and
\IEEEauthorblockN{Tim Rehbronn\textsuperscript{\textdagger\orcidicon{0009-0009-5241-2588}}}
\IEEEauthorblockA{\textit{Institute for Automotive Engineering} \\
\textit{RWTH Aachen University}\\
Aachen, Germany \\
tim.rehbronn@rwth-aachen.de}
\and
\IEEEauthorblockN{Abin Jose\textsuperscript{\orcidicon{0000-0002-3974-3552}}}
\IEEEauthorblockA{\textit{Institute of Imaging and Computer Vision} \\
\textit{RWTH Aachen University}\\
Aachen, Germany \\
abin.jose@lfb.rwth-aachen.de}
\and
\IEEEauthorblockN{Johannes Stegmaier\textsuperscript{\orcidicon{0000-0003-4072-3759}}}
\IEEEauthorblockA{\textit{Institute of Imaging and Computer Vision} \\
\textit{RWTH Aachen University}\\
Aachen, Germany \\
johannes.stegmaier@lfb.rwth-aachen.de}
\and
\IEEEauthorblockN{Bastian Lampe\textsuperscript{\orcidicon{0000-0002-4414-6947}}}
\IEEEauthorblockA{\textit{Institute for Automotive Engineering} \\
\textit{RWTH Aachen University}\\
Aachen, Germany \\
bastian.lampe@ika.rwth-aachen.de}
\and
\IEEEauthorblockN{Timo Woopen\textsuperscript{\orcidicon{0000-0002-7177-181X}}}
\IEEEauthorblockA{\textit{Institute for Automotive Engineering} \\
\textit{RWTH Aachen University}\\
Aachen, Germany \\
timo.woopen@ika.rwth-aachen.de}
\linebreakand
\IEEEauthorblockN{Lutz Eckstein}
\IEEEauthorblockA{\textit{Institute for Automotive Engineering} \\
\textit{RWTH Aachen University}\\
Aachen, Germany \\
lutz.eckstein@ika.rwth-aachen.de}
}

\maketitle
\conf{\textit{  III. International Conference on Electrical, Computer and Energy Technologies (ICECET 2023) \\ 
16-17 November 2023, Cape Town-South Africa}}

\copyrightnotice

\begin{abstract}
Automated vehicles require an accurate perception of their surroundings for safe and efficient driving. Lidar-based object detection is a widely used method for environment perception, but its performance is significantly affected by adverse weather conditions such as rain and fog. In this work, we investigate various strategies for enhancing the robustness of lidar-based object detection by processing sequential data samples generated by lidar sensors. Our approaches leverage temporal information to improve a lidar object detection model, without the need for additional filtering or pre-processing steps. We compare $10$ different neural network architectures that process point cloud sequences including a novel augmentation strategy introducing a temporal offset between frames of a sequence during training and evaluate the effectiveness of all strategies on lidar point clouds under adverse weather conditions through experiments. Our research provides a comprehensive study of effective methods for mitigating the effects of adverse weather on the reliability of lidar-based object detection using sequential data that are evaluated using public datasets such as nuScenes, Dense, and the Canadian Adverse Driving Conditions Dataset. Our findings demonstrate that our novel method, involving temporal offset augmentation through randomized frame skipping in sequences, enhances object detection accuracy compared to both the baseline model (Pillar-based Object Detection) and no augmentation.
\end{abstract}



\section{Introduction}
Automated driving in adverse weather conditions requires robust object detection. Object detection models trained on particular datasets display a bias towards the dataset-specific characteristics such as the weather conditions represented in the training data. As a consequence, these models tend to perform well in similar weather conditions but experience a notable decline in performance when confronted with substantially different weather conditions not present in the training data. Transductive Transfer Learning pertains to the Domain adaptation approach wherein the task, specifically object detection in our case, remains consistent between the target and source domains, while the marginal distribution of data diverges due to distinct weather conditions~\cite{domain_adaption_overview}. Recent approaches to improving this domain adaptation to adverse weather conditions include e.g. data pre-processing and fusing detections from multiple sensors. However, improving the performance of each detector itself will increase the robustness of the overall perception system.

Another way to enhance the robustness of lidar-based object detection under adverse weather conditions is the use of time series data as input, as lidar sensors capture sequences of point clouds. Data augmentation is a common way to increase the robustness of a trained model by slightly modifying existing data. This work presents a novel augmentation strategy for using temporal information within a sequence of lidar point clouds by adding a random temporal offset, i.e. modifying the temporal distance between frames in a data sequence. Our contributions include a novel architecture based on Pillar-based Object Detection~\cite{PBOD}, which allows using temporal information of sequences, a comprehensive study of different ways for using data sequences in the model architecture, and a quantitative evaluation and comparison of the trained models on three real-world adverse weather datasets.

\section{Background}

State-of-the-art neural networks such as TANet~\cite{liu2019tanet} demonstrate robust object detection on simulated noisy point clouds, indicating their potential to perform well under real adverse weather conditions. Recent datasets, including adverse weather conditions such as Dense~\cite{gated2depth2019} and Canadian Adverse Driving Conditions~\cite{cadc}, have been published to train these neural networks. Filtering methods such as Dynamic Statistical Outlier Removal (DSOR)~\cite{DSOR} can be applied in a pre-processing stage while training and inference to de-rain or defog sparse point clouds, but it requires an additional calculation step in the model's data pipeline.

Lidar sensors are active, remote-sensing devices that emit laser pulses and measure the time-of-flight of the reflected beams to calculate distances~\cite{IEEE_sensors}. They are capable of generating a sparse representation of the environment with accurate depth information. However, when adverse weather such as fog, mist, rain, or snow occurs, the laser beams emitted by the lidar sensors have to travel through a volume of distorting particles. The presence of scattering particles in adverse weather conditions can induce diffraction or absorption of laser beams, resulting in attenuated or absent reflections, leading to missed detections of objects of interest, such as cars~\cite{adverse_weather_physics}. Moreover, these particles may cause backscattering of the laser beam, introducing undesirable reflections and noise in the point cloud data~\cite{adverse_weather_physics}. Both missed detections and undesired reflections contribute to significant degradation in the performance of lidar sensors. Consequently, object detection, scene understanding, and safe navigation may not be ensured.

Adverse weather can be classified into two categories: static and dynamic. Static adverse weather includes fog, mist, or haze, which are usually present in a stable form over a relatively long period of time. On the other hand, dynamic adverse weather, such as rain or snow, occurs in a more unpredictable and changing manner. In rainy or snowy weather, the laser beams emitted by the sensor interact with hydro-meteors like water that occur densely in the air~\cite{adverse_weather_2}.

\begin{figure}[hbt]
    \centering
    \includegraphics[width=0.5\textwidth]{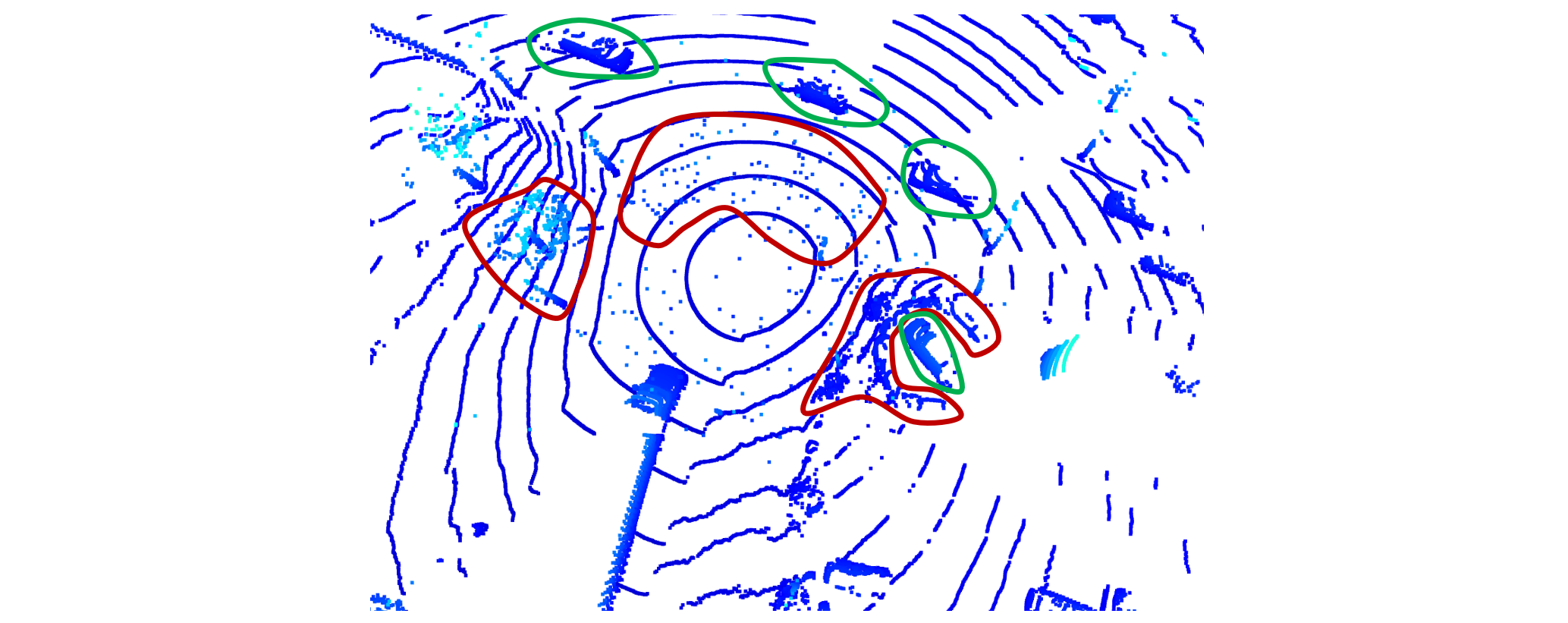}
    \caption{Noise in a point cloud; red markers illustrate noisy points, and green markers show points belonging to objects. Shades of blue indicate reflection intensity. Point Cloud from Canadian Adverse Driving Conditions dataset~\cite{cadc}.}
    \label{fig:noise}
\end{figure}

As illustrated in Figure \ref{fig:noise}, the point cloud representation of the environment becomes noisy when adverse weather occurs, with particles having a high reflection leading to many undesired reflections~\cite{adverse_weather_1}. This can cause limited visibility range, object detector degradation due to noise degradation, and objects blocked by reflecting particles~\cite{adverse_weather_3}. Improving the robustness of lidar-based object detection in adverse weather conditions to mitigate the described sensor degradations is a crucial task to enable automated driving in such scenarios.

\section{DATASETS AND RELATED WORK}

This section discusses the datasets used in this work and related methods proposed to increase the robustness of lidar-based object detection.

\subsection{Datasets}
For object detection algorithms to work effectively under adverse weather conditions, it's essential to include these conditions in the data to train the detector to handle noisy data.

\subsubsection{DENSE}
The Dense dataset~\cite{gated2depth2019} captures $10,000$ km of driving through Germany, Denmark, Sweden, and Finland with over $100,000$ annotated objects with 3D and 2D information in $13,770$ samples. It contains various weather conditions like rain, fog, snow, and clear weather. This work uses a subset of the dataset comprising $12,930$ frames ignoring frames that do not form valid data sequences. The validation set consists of every tenth sample of the original set. The dataset is reduced to ensure that only valid data sequences are contained. Therefore, the training set contains $11,639$ samples and the test set contains $281$ samples. The dataset annotates boxes up to $80$ m of distance, but since the annotation process uses front-view camera images to position the bounding boxes correctly, only labels in front of the car are available.

\subsubsection{Canadian Adverse Driving Conditions}
The Canadian Adverse Driving Conditions dataset~\cite{cadc} is a multi-modal dataset that includes lidar and camera data captured in December 2019 in Waterloo and provides data with snowy weather conditions. It contains $7,000$ labeled samples with various snowfall conditions. Samples without objects are removed, leaving the training set with $6,249$ samples and the validation set with $649$ samples.

\subsubsection{nuScenes}
The nuScenes dataset~\cite{nuscenes_paper} is a multi-modal dataset comprising six cameras, five radars, and one lidar sensor. It provides $40,000$ lidar samples with 3D object information captured in Boston and Singapore at day and night time as well as rainy conditions. The training set of the dataset contains $28,130$ samples and the validation set contains $6,019$ samples ensuring that only valid data sequences are used.

\subsection{Camera-based Object Detection}
Image-based detection predicts 3D bounding box and classifies objects from 2D RGB image data. Fast R-CNN~\cite{fast_RCNN} uses selective search to generate region proposals, while Faster R-CNN~\cite{faster_RCNN} omits selective search and applies an RPN to increase efficiency. Stereo R-CNN~\cite{stereo_RCNN} produces 3D object proposals in each image and aligns proposals of images to refine the prediction result. SMOKE~\cite{SMOKE} predicts center coordinates of the 3D bounding box projected on a 2D image plane and regression to obtain bounding box parameters. Other methods estimate depth images and use them to generate lidar pseudo point clouds that can be used for any lidar-based object detection network~\cite{Pseudo_lidar}.

\subsection{Lidar-based Object Detection}
Conventional models for lidar-based object detection operate on raw point clouds like PointNet~\cite{PointNet} and PointNet++\cite{PointNet++}, or ordered point clouds like VoxelNet\cite{VoxelNet} or PointPillar~\cite{PointPillar}. Many methods project the point clouds on different planes to retain more information in the abstraction process~\cite{OD_Overview_IEEE}. The PBOD model~\cite{PBOD} is used as a baseline state-of-the-art object detector for the experiments in this work. PBOD extends the idea of Point Pillar by using an additional view projection like in Multi-View Fusion (MVF)~\cite{MultiView_PBOD}, and avoids predictions per anchor by predicting bounding boxes per pillar. PBOD also uses a cylindrical view projection, resulting in a less distorted projection than the spherical projection used in MVF.

\subsection{Sequential Data Processing}

Various methods use sequential point clouds to enhance object detection, such as reusing bounding box center predictions of previous frames, concatenating features of point clouds, or using a pillar messaging network to pass information between pillars of a frame and encode information between time steps of point clouds via a spatio-temporal attention transformer module with a convolution gated-recurrent-unit recurrent network (convGRU)~\cite{LSTM_lidar_OD, temporal_lidar_Trafo_GRU}. MPPnet~\cite{MPPnet} uses proxy points to integrate multi-frame features from a sequence of point clouds.

Convolutional Long Short-Term Memory (convLSTM) is used in some approaches to fuse information from preceding lidar frames with information from the current frame. In~\cite{McCrae2020}, the authors introduce a convLSTM layer in the PointPillars~\cite{PointPillar} architecture to reduce the number of samples in one sequence to $3$ while still outperforming the previous approach using $10$ samples. YOLO4D~\cite{YOLO4D} extends YOLO3D~\cite{yolo3d} with either frame stacking or a conLSTM layer and shows that both improve the performance and robustness of the trained model.

Another stream of research for sequential data processing is the transformer, a neural network originating from natural language processing~\cite{attention_is_all_you_need} that can efficiently aggregate information of the whole image or point cloud with relations of each point to every other point in the point cloud or image and thus also has the potential for object detection. Transformers are used, e.g., to fuse camera and lidar data efficiently~\cite{LIFT} or directly perform object detection on camera images~\cite{DETR} or large lidar point clouds~\cite{Transformer_Lidar, Transformer_Lidar_2}.

\section{METHODOLOGY}

This work aims to answer the following research questions:

\begin{enumerate}
\item How well does the lidar-based object detector, Pillar-Based Object Detection~\cite{PBOD}, perform when faced with noisy data collected in adverse weather conditions?
\item Is the use of sequential lidar data capable of improving the object detection model? 
\item In which ways can augmentation of sequential lidar data improve the object detection model? 
\item How do our approaches using temporal information compare to other state-of-the-art methods?
\end{enumerate}

\begin{figure}[tb]
    \centering
    \smallskip\smallskip
    \includegraphics[width=0.45\textwidth]{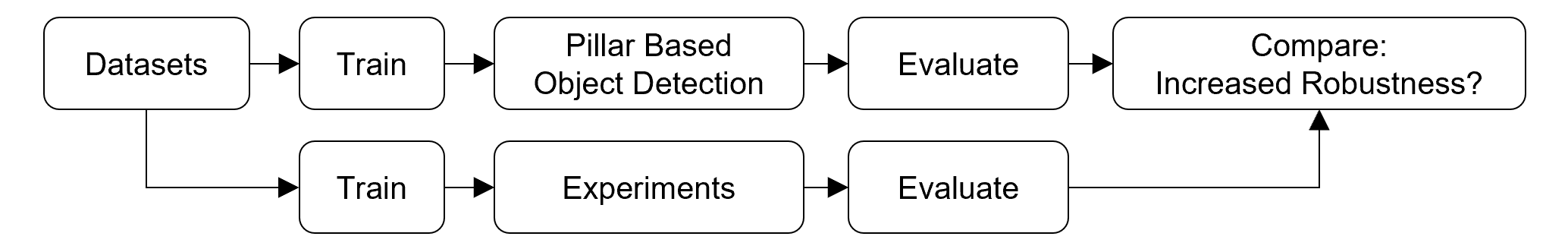}
    \caption{We compare $10$ different neural network architectures using temporal data sequences and used all architectures to train models on $3$ datasets. The models are evaluated and compared to a baseline model based on Pillar-based Object Detection~\cite{PBOD} on validation splits of these datasets.}
    \label{fig:experiment_overview}
\end{figure}

To address these research questions, we extend the object detection model using various approaches to leverage sequential lidar point clouds. We also use other state-of-the-art techniques for robust object detection in point clouds and compare their performance against the baseline PBOD model. An overview of our experimental approach is shown in Figure~\ref{fig:experiment_overview}. To evaluate and compare the models, we use the mean average precision at an IoU threshold of $0.5$ and $0.75$, as well as the inference time required for a single sequence.

This study conducted four experiments on point cloud sequences, utilizing temporal information. The trainings and evaluations were performed utilizing an NVIDIA A100 GPU equipped with 40 GB of VRAM. To cope with hardware limitations, a filter factor (FF) can be used to reduce the number of adjustable parameters in the model. This factor was set to $1$ initially and the batch size was set as large as possible on the given hardware.

\begin{figure}[hbt!]
    \centering
    \includegraphics[width=0.45\textwidth]{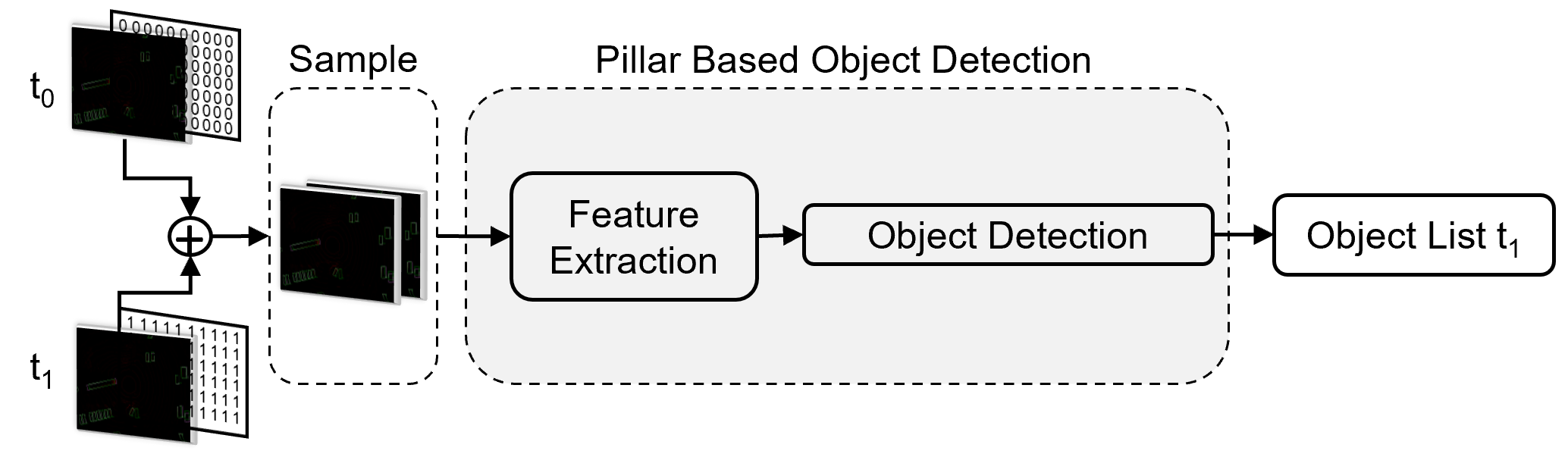}
    \caption{IC - We concatenate temporally succeeding point clouds in the input stage to fuse the information about the current driving scene contained in both point clouds, i.e. concatenating points from two sequential point clouds.}
    \label{fig:Experiments_1_2}
\end{figure}

The first two experiments are Input Concatenation (IC) and Input Concatenation with temporal encoding (IC+), as shown in Figure~\ref{fig:Experiments_1_2}. IC concatenates the points of multiple point clouds from different time steps to form one common point cloud to increase density, e.g. for two point clouds $P_1$ and $P_2$ consisting of points $p$:

\begin{eqnarray}
    p & = & (x, y, z, intensity) \\
    P_1 & = & \{ p_i | 1 \le i \le m \} \\
    P_2 & = & \{ p_i | 1 \le i \le n \} \\
    P_{IC} & = & \{ p_1, ..., p_m, p_{m+1}, ..., p_n \}
\end{eqnarray}

Furthermore, IC+ adds another attribute to each point indicating the temporal position of the point cloud in the sequence: 

\begin{equation}
p = (x, y, z, intensity, \{ 0,1 \})    
\end{equation}

\begin{figure}[hbt]
    \centering
    \smallskip\smallskip
    \includegraphics[width=0.45\textwidth]{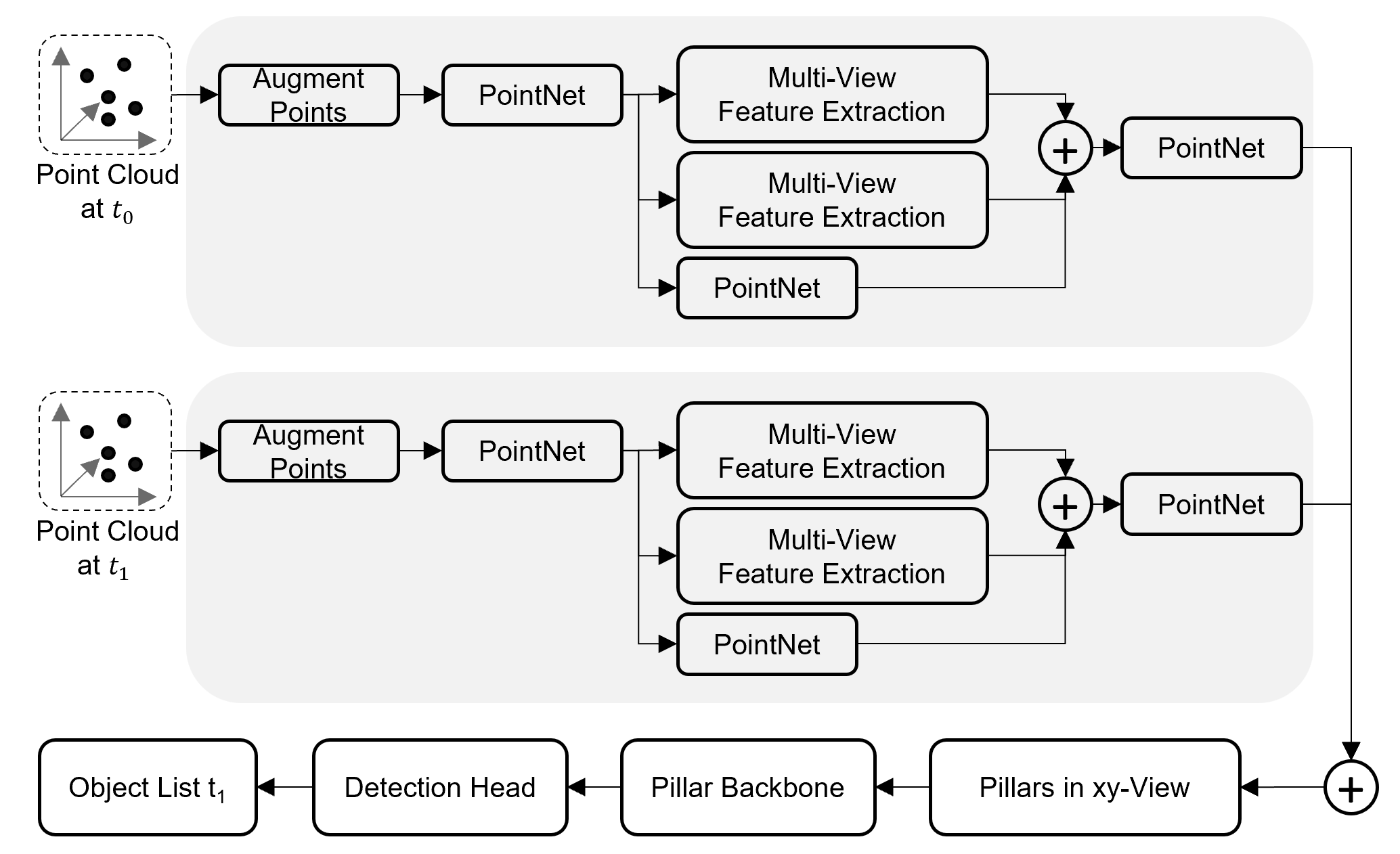}
    \caption{FC - We fuse sequential samples later in the network after features are extracted by feature concatenation.}
    \label{fig:Experiments_3}
\end{figure}

The third experiment, Feature Concatenation (FC), is displayed in Figure~\ref{fig:Experiments_3}. FC fuses data later in the detection pipeline by creating a separate feature extraction branch for each frame in the input sequence, which is concatenated before the pillars are projected back to a birds-eye-view. In another experiment FC+, a multi-layer perceptron is added after feature concatenation to increase the model complexity.

\begin{figure}[hbt]
    \centering
    \includegraphics[width=0.45\textwidth]{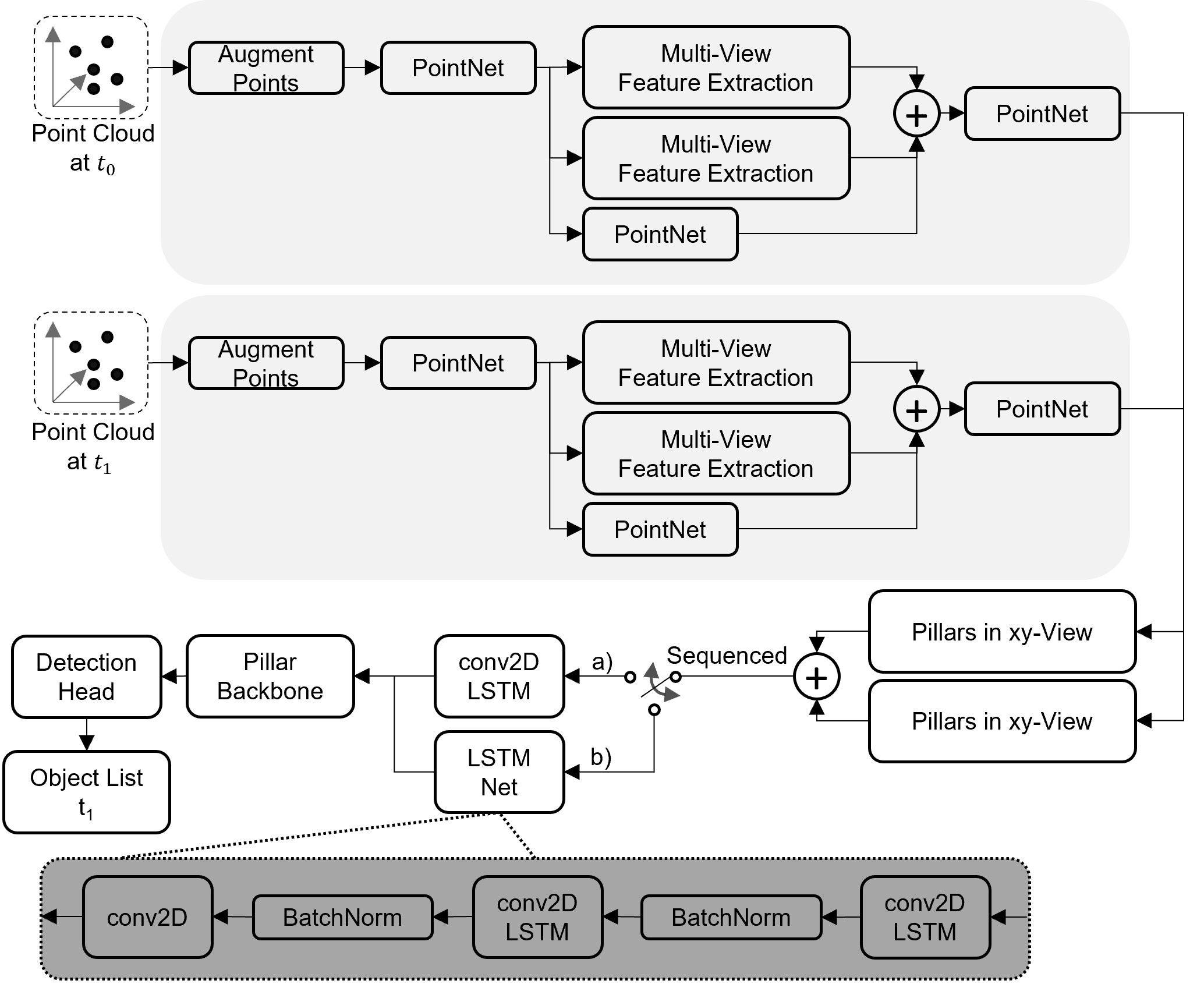}
    \caption{LSTM - We fuse sequential samples later in the network by building a memory with extracted features using long short-term memory cells.}
    \label{fig:Experiments_4}
\end{figure}

The fourth experiment, Long Short-Term Memory (LSTM), is depicted in Figure~\ref{fig:Experiments_4}. LSTM separately extracts features of each point cloud in the sequence, building pseudo-images for each point cloud, which are input to a convLSTM module that calculates a feature map for further processing in the PBOD network. Fused information is fed into the pillar backbone of the default PBOD model, generating even more discriminative features. Additionally, the single convLSTM is extended to a network of convLSTM cells to increase learning capability.

\begin{figure}[hbt]
    \centering
    \includegraphics[width=0.45\textwidth]{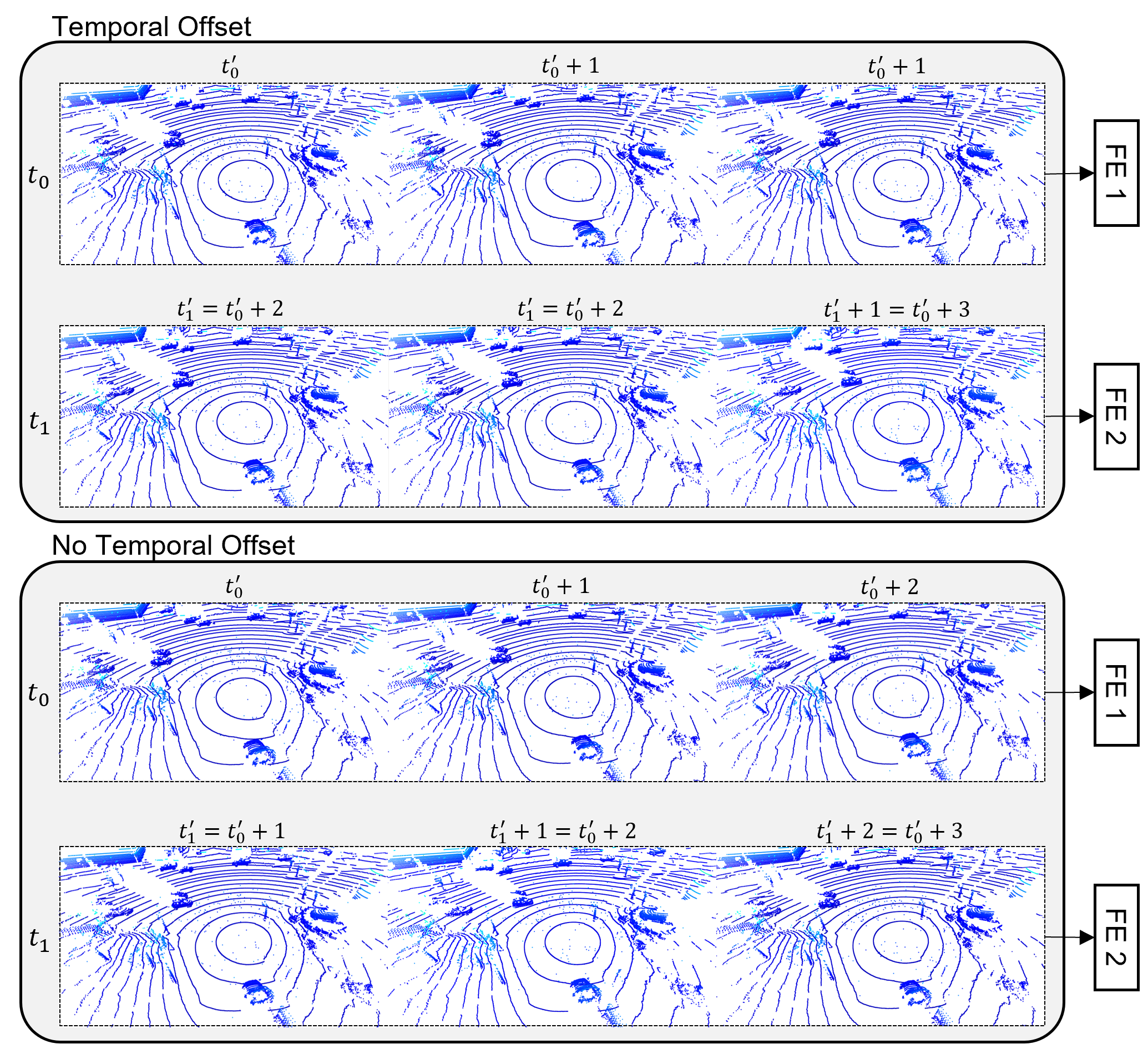}
    \caption{Consecutive frames in a sequence with and without temporal offset. The upper and lower two rows each represent one batch of input samples with a batch size of $3$, while $t_0$ is the first and $t_1$ is the second element of one sample with a sequence length of $2$. $t'_i$ indicates the timestamp when the point cloud was recorded and FE is the feature extraction module.}
    \label{fig:temporal_offset}
\end{figure}

In addition, this study investigates the effects of introducing a temporal offset between consecutive frames in a sequence, specifically a skipping of frames, as depicted in Figure~\ref{fig:temporal_offset}. This means that the time interval between samples in a sequence is varied randomly during training. The upper part of the figure introduces a temporal offset for two of the three frames in the sequence, while one is connected to the preceding frame. The lower part displays the processing of consecutive lidar frames in a sequence to maintain the original temporal relationship of the point cloud stream, preserving the temporal structure even when shuffling is enabled.

\section{RESULTS AND DISCUSSION}

\begin{table*}[!htb]
\caption{Evaluation results. Inference Time in $\frac{\text{seconds}}{\text{iteration}}$; mAP at $\text{IoU} = \{0.5,0.75\}$; (*) with temporal offset; \\BS=~batch size; FF=~filter factor; SQ=~sequence length.}
\begin{center}
\resizebox{0.85\linewidth}{!}{%
\begin{tabular}{ |p{5cm}|p{2cm}|p{2cm}|p{2cm}|p{2cm}|}
 \hline
 \multicolumn{5}{|c|}{Overall Evaluation Results} \\
 \hline
 Model (BS/FF/SQ) & Dataset & sec/it & mAP (0.5)& mAP (0.75)\\
 \hline\hline
  PBOD (8/1/-) & Dense & \textbf{0.096} & 0.567 & 0.112 \\ 
 \hline
  IC  & Dense & 0.101 & 0.015 & 0.002 \\  
 \hline
  IC+   & Dense & 0.094 & 0.005 & 0.000 \\
 \hline
  FC (4/1/2) & Dense & 0.164 & 0.000 & 0.000 \\
 \hline
  FC+ (4/1/2) & Dense & 0.163 & 0.000 & 0.000 \\
 \hline
 FC* (4/1/2) & Dense & 0.161 & \textbf{0.590} & \textbf{0.141} \\
 \hline
  FC+* (4/1/2) & Dense & 0.164 & 0.579 & 0.125 \\ 
 \hline
  LSTM (2/1/2) & Dense & 0.171 & 0.488 & 0.138 \\
 \hline
  LSTM Net (2/1/2) & Dense & 0.177 & 0.419 & 0.134 \\
 \hline
  LSTM* (6/2/2) & Dense & 0.156 & 0.583 & 0.106 \\
  \hline
  LSTM Net* (4/2/2) & Dense & 0.160 & 0.560 & 0.138 \\
 \hline\hline
  PBOD (8/1/-) & CADC & \textbf{0.092} & 0.641 & 0.313 \\ 
 \hline
  IC & CADC & 0.100 & 0.513 & 0.241 \\ 
 \hline
  IC+ & CADC & 0.097 & 0.508 & 0.228 \\
 \hline
  FC (8/2/2) & CADC & 0.140 & 0.000 & 0.000 \\ 
 \hline
  FC+ (4/1/2) & CADC & 0.140 & 0.000 & 0.000 \\
 \hline
  FC* (6/1/2) & CADC & 0.168 & 0.672 & 0.315 \\ 
 \hline
  FC+* (4/1/2) & CADC & 0.162 & 0.643 & 0.343 \\ 
 \hline
  LSTM (2/1/2) & CADC & 0.173 & 0.000 & 0.000 \\
 \hline
  LSTM Net (2/1/2) & CADC & 0.177 & 0.008 & 0.006 \\
 \hline
  LSTM* (6/1/2) & CADC & 0.164 & \textbf{0.680} & 0.340 \\
 \hline
  LSTM Net* (4/2/2) & CADC & 0.202 & 0.653 & \textbf{0.348} \\
 \hline\hline
  PBOD (8/1/-) & NuScenes & \textbf{0.099} & 0.510 & 0.246\\
 \hline
  IC & NuScenes & 0.109 & 0.387 & 0.144 \\ 
 \hline
  IC+ & NuScenes & 0.106 & 0.270 & 0.100 \\
 \hline
  FC (4/1/2) & NuScenes & 0.168 & 0.000 & 0.000 \\
 \hline
  FC+ (4/1/2) & NuScenes & 0.166 & 0.000 & 0.000 \\
 \hline
  FC* (4/1/2) & NuScenes & 0.166 & 0.474 & 0.246 \\
 \hline
  FC+* (4/1/2) & NuScenes & 0.162 & 0.558 & \textbf{0.266} \\
 \hline
 LSTM* (6/1/2) & NuScenes & 0.169 & \textbf{0.623} & 0.200 \\
 \hline
  LSTM Net* (6/1/2) & NuScenes & 0.184 & 0.349 & 0.143 \\
 \hline
\end{tabular}
}
\label{tab:Overall_Evaluation_Results}
\end{center}
\vspace{-7pt}
\end{table*}

Table~\ref{tab:Overall_Evaluation_Results} shows the evaluation results of our experiments. It is apparent that Input Concatenation (IC, IC+) with two samples only worsens the results. Furthermore, Feature Concatenation without temporal offset, i.e. FC and FC+, does not lead to any correct predictions for all datasets. However, introducing a temporal offset in FC* and FC+* significantly improves the results and FC* yielded the best evaluation results on the Dense dataset, with a mean average precision of $0.590$ and $0.141$ at IoU of $0.5$ and $0.75$, respectively. This increased the performance by about $4.1\%$ at IoU of $0.5$ and by about $25.9\%$ at IoU of $0.75$, compared to the default PBOD model. However, this also led to an increase in inference time by about $68\%$, to $0.16$ seconds per input sequence.

For the CADC dataset, the model based on the PBOD with an additional convLSTM module before the pillar backbone performed best at IoU of $0.5$ with mAP of $0.680$ ($+6.1\%$) and $0.340$ ($+8.6\%$). Using a convLSTM network improved detection results at IoU of $0.75$ by about $11.2\%$, making the complex model more confident in correct predictions than the single convLSTM cell. However, the inference time further increased with $0.202$ compared to $0.164$ seconds per iteration.

Finally, the modified PBOD model with a convLSTM module and a random temporal offset performed best on the nuScenes dataset, with a $22.2\%$ increase in mAP at IoU of $0.5$, but a $18.7\%$ decrease at IoU of $0.75$. At IoU of $0.75$, the FC experiment with random time offset performed best, increasing mAP by $8.1\%$. All other models showed inferior detection performance on all datasets compared to the default Pillar-based Object Detection model. 

Potential reasons for the observed performance improvement using a randomized skipping of frames may stem from the model's ability to adopt a general approach to utilize temporal information rather than learning the underlying temporal patterns specific to the dataset. In the absence of a randomized temporal offset, the model tends to learn the inconsistent underlying patterns, which are present in the dataset, leading to degraded object detection performance.

Furthermore, the experiments involving input concatenation might be affected by the absence of coordinate transformation between successive point clouds due to the lack of temporal information in the framework. This early-stage concatenation with a missing abstraction of information, as evident in feature concatenation, could result in making the model more susceptible to misalignments in the overlapping point cloud data.

Moreover, the dataset comprises samples with substantial temporal distances, resulting in chaotic overlapping point clouds, as each time step's point clouds describe distinct driving scenes. These factors can contribute to the observed effects on the model's performance.

\section{CONCLUSION AND OUTLOOK}

In conclusion, the experiments conducted demonstrate that introducing a random temporal offset between frames in a sequence can enhance detection performance on all three state-of-the-art datasets. This approach makes the model more robust against noisy data and improves detection results in adverse weather conditions. The observed performance improvement is notable, even in the nuScenes dataset, which mainly contains clear weather data, indicating the model's enhanced generalization ability. While input concatenation did not yield better performance, feature concatenation improved detection results. Moreover, adding a convLSTM module to the network significantly increased detection performance on all datasets compared to the vanilla PBOD model, whereas adding more layers to the network did not improve the performance. Future research should investigate how many frames should be used in a sequence and also explore the use of Transformer architecture to improve detection performance by leveraging the temporal information of frames and improving inference time, as current works show promising results.

\section*{ACKNOWLEDGEMENT}

This research is accomplished within the project ”AUTOtech.agil” (FKZ 01IS22088A). We acknowledge the financial support for the project by the Federal Ministry of Education and Research of Germany (BMBF).


\bibliographystyle{IEEEtran}
\bibliography{literature}


\end{document}